\documentclass[10pt,twocolumn,letterpaper]{article}

\usepackage{cvpr}              

\definecolor{cvprblue}{rgb}{0.21,0.49,0.74}
\usepackage[pagebackref,breaklinks,colorlinks,allcolors=cvprblue]{hyperref}
\usepackage{amssymb}
\usepackage{xcolor}
\usepackage{multirow}
\usepackage{colortbl}
\usepackage{float}
\def\paperID{16007} 
\def\confName{CVPR}
\def\confYear{2025}

\title{ANNEXE: Unified \underline{An}alyzing, A\underline{n}sw\underline{e}ring, and Pi\underline{x}el Grounding \\
	 for \underline{E}gocentric Interaction}

\author{Yuejiao Su, Yi Wang\textsuperscript{\footnotemark[1]}, Qiongyang Hu, Chuang Yang, Lap-Pui Chau\textsuperscript{\footnotemark[1]}\\
Department of Electrical and Electronic Engineering, \\
The Hong Kong Polytechnic University, Hong Kong SAR\\
{\tt\small \{yuejiao.su,qiongyang.hu\}@connect.polyu.hk,\{yi-eie.wang,c1yang,lap-pui.chau\}@polyu.edu.hk}\\
\href{https://yuggiehk.github.io/annexe/}{https://yuggiehk.github.io/annexe/}}

\begin{document}
\maketitle
\footnotetext[1]{Corresponding author.}
\begin{abstract}
Egocentric interaction perception is one of the essential branches in investigating human-environment interaction, which lays the basis for developing next-generation intelligent systems. However, existing egocentric interaction understanding methods cannot yield coherent textual and pixel-level responses simultaneously according to user queries, which lacks flexibility for varying downstream application requirements. To comprehend egocentric interactions exhaustively, this paper presents a novel task named Egocentric Interaction Reasoning and pixel Grounding (Ego-IRG). Taking an egocentric image with the query as input, Ego-IRG is the first task that aims to resolve the interactions through three crucial steps: analyzing, answering, and pixel grounding, which results in fluent textual and fine-grained pixel-level responses.
Another challenge is that existing datasets cannot meet the conditions for the Ego-IRG task. To address this limitation, this paper creates the Ego-IRGBench dataset based on extensive manual efforts, which includes over 20k egocentric images with 1.6 million queries and corresponding multimodal responses about interactions. Moreover, we design a unified ANNEXE model to generate text- and pixel-level outputs utilizing multimodal large language models, which enables a comprehensive interpretation of egocentric interactions.
The experiments on the Ego-IRGBench exhibit the effectiveness of our ANNEXE model compared with other works. 
\end{abstract}    
\vspace{-0.7cm}
\section{Introduction}
\label{sec:intro}


\begin{figure}[t]
	\centering
	\includegraphics[width=\linewidth]{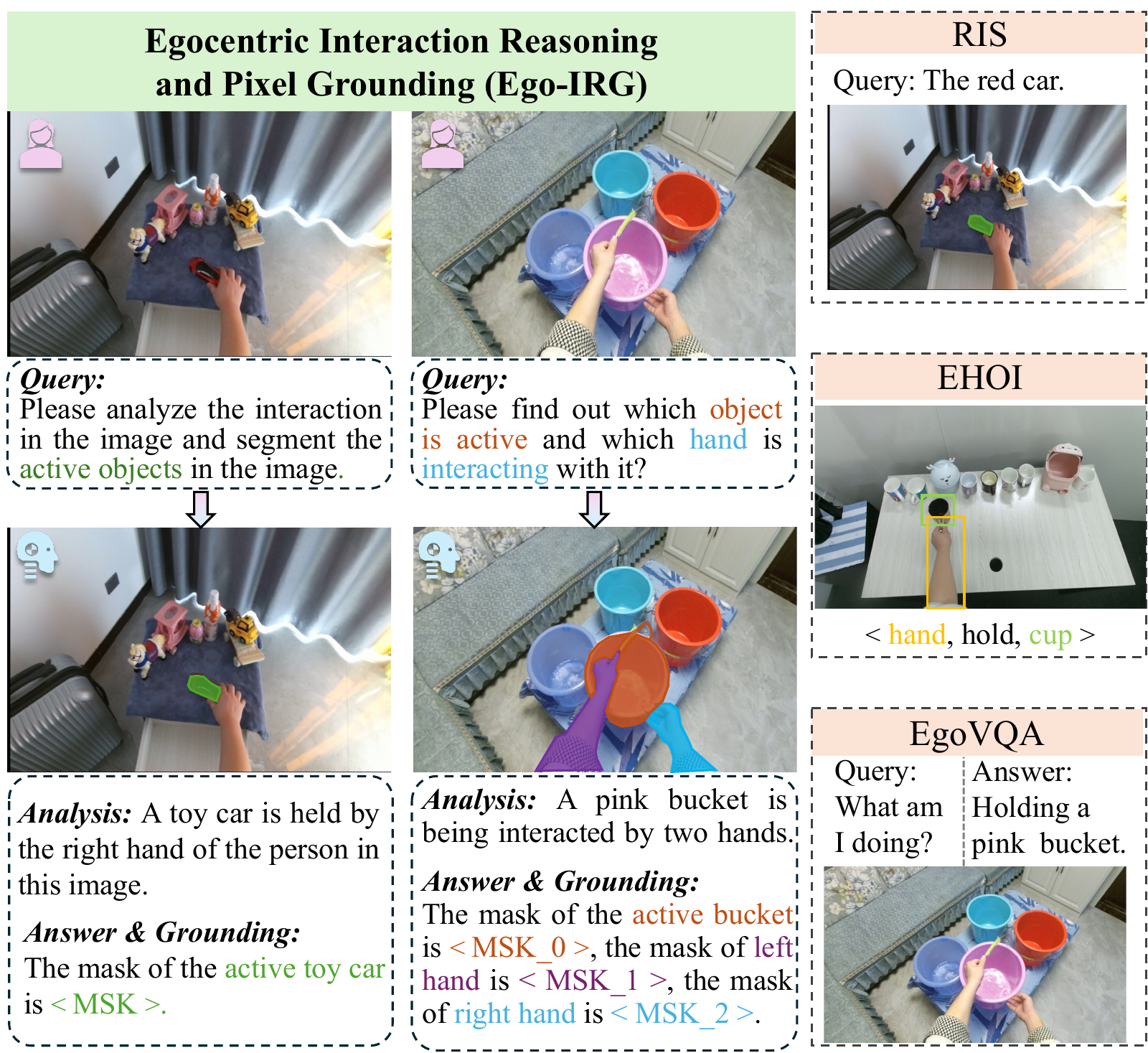}
\caption{Ego-IRG task versus other related tasks. The Ego-IRG task advances the interaction understanding compared with other related tasks such as Referring Image Segmentation (RIS), Egocentric Hand-Object Interaction detection (EHOI), and Egocentric Question Answering (EgoVQA). }
	\label{fig:1}
\vspace{-0.8cm}
\end{figure}

Understanding how humans interact with the environment or objects through visual information has always been a pivotal focus in computer vision, which can contribute to constructing next-generation agent systems.
Traditionally, studies in this area have concentrated on exocentric or third-person view (TPV) interaction analysis, achieving substantial progress through evolving various tasks such as human-object interaction detection (HOI-det) \cite{DBLP:conf/cvpr/LeiYL24,ning2023hoiclip} and action recognition \cite{zhang2024pevl,chen2024align}.
Recently, a growing number of users spontaneously post videos and images taken from egocentric or first-person view (FPV) \cite{li2024egogen} on social media platforms due to the development of head-mounted devices (HMD) and GoPro.
These egocentric data provide a unique perspective by showing visual signals from the perspective of the camera wearer, emphasizing individual behavior in different situations. The study of hand-object interaction from an egocentric perspective can gain deeper insights into human interaction, which has a wide range of applications such as augmented reality (AR) \cite{chalasani2018egocentric}, embodied AI \cite{suglia2024alanavlm}, and industrial human-assistive systems \cite{leonardi2024exploiting}.

However, severe occlusions and background variations pose significant challenges to FPV interaction study and create a huge domain gap between TPV and FPV, making it non-trivial to directly transfer exocentric algorithms to egocentric interaction analysis \cite{li2022egocentric}.
With the emergence of various large-scale egocentric datasets such as Ego4D \cite{grauman2022ego4d}, HOI4D \cite{liu2022hoi4d}, and EPIC-KITCHENS \cite{Damen2022RESCALING,VISOR2022}, some recent works utilized egocentric visual information to detect \cite{leonardi2025synthetic}, segment \cite{zhang2022fine,su2024ormnet}, and recognize the interactions \cite{wang2023ego} between hands and objects. These studies aim to generate bounding boxes or masks for hands and objects while classifying action categories.
Parallel to this, other research efforts have focused on producing egocentric video narrations \cite{shen2024learning,xu2024retrieval}, which provide the text responses to describe the sequential interaction in detail.
Although these advancements have significantly contributed to understanding human-object interaction from an egocentric perspective, few studies have successfully integrated coherent text-level and fine-grained pixel-level responses as outputs.
These multi-modal outputs are all essential for a thorough comprehension of egocentric interactions.
Furthermore, the responses of existing works are in fixed mode, which lacks flexibility when egocentric interaction results are employed for diverse downstream tasks.

In this paper, we propose a novel task called \textbf{Ego}centric \textbf{I}nteraction \textbf{R}easoning and pixel-\textbf{G}rounding (\textbf{Ego-IRG}) to facilitate the study of comprehensive egocentric interaction. As shown in Fig. \ref{fig:1}, this task allows for unified analyzing, answering, and pixel grounding regarding interactions within egocentric images based on various user queries.
Specifically, given a text query and an egocentric image, the Ego-IRG task encompasses two essential steps for comprehensive interaction parsing: reasoning and pixel grounding. 
The reasoning step aims to generate the text-level responses for interaction interpretation, which includes two sub-tasks: analyzing and answering.
The analyzing sub-task aims to generate an overall text description of interactions depicted in the image, while the answering sub-task is to reply to user queries that the specific objects need to be segmented.
Furthermore, the pixel grounding step aims to locate objects relevant to the query using fine-grained masks.
The challenging Ego-IRG task unifies several egocentric vision tasks that are commonly addressed separately, such as referring image segmentation (RIS) \cite{xia2024gsva,liu2023gres}, egocentric hand-object interaction detection (EHOI) \cite{leonardi2025synthetic}, and egocentric visual question answer (EgoVQA) \cite{di2024grounded}.
By exploring this novel Ego-IRG task, egocentric interactions can be studied more systematically and query-oriented, resulting in flexibility when applied to real-world applications.

Nevertheless, the existing datasets cannot meet the requirements for the Ego-IRG task. To address this issue and facilitate subsequent studies, this paper constructs a large-scale dataset, \textbf{Ego}centric \textbf{I}nteraction \textbf{R}easoning and \textbf{G}rounding \textbf{Bench}mark (\textbf{Ego-IRGBench}).
Built upon the HOI4D \cite{liu2022hoi4d} dataset, we employed extensive manual efforts to relabel egocentric images with interaction descriptions and query-answer-mask pairs.
The Ego-IRGBench comprises 20,681 egocentric images, each accompanied by descriptions depicting the interactions. In addition, over 1.6 million query-answer pairs, along with pixel-level grounding results, are included in the Ego-IRGBench. 

Moreover, few studies have utilized depth information to enhance understanding of egocentric interactions.
We argue that depth information is essential and helpful to distinguish the active objects. 
Specifically, when humans perform an action, they typically position the object of interest in the foreground, creating a notable depth distinction between hands and manipulating objects (foreground) relative to the background. 
The depth differences can be utilized to recognize the interacting objects, especially in scenarios where multiple objects of similar appearance and categories exist.
Therefore, this paper introduces a unified depth-assisted \textbf{AN}alyzing, a\textbf{N}sw\textbf{E}ring, and pi\textbf{X}el grounding for \textbf{E}gocentric interaction (\textbf{ANNEXE}) network to tackle the novel Ego-IRG task. 
This network incorporates the text generation module and the mask generation module to tackle the reasoning and pixel grounding simultaneously.
In addition, due to the success of multimodal large language models (MLLMs), we utilize MLLMs in the reasoning module to enhance the representations of queries and images.
To summarize, our work has three main contributions:
\begin{itemize}
	\item To facilitate the study of human behavior, we present the new Ego-IRG task to interpret egocentric interaction comprehensively by a synergy of three ego-tasks: analyzing, answering, and pixel grounding.
    \item To address benchmark limitations, we propose a large-scale annotated Ego-IRGBench dataset containing interaction descriptions for over 20k egocentric images and 1.6M query-answer-mask paired labels.
	\item We present the ANNEXE for tackling the Ego-IRG task utilizing MLLMs, which is the first model that can understand visual-language inputs and generate text- and pixel-level responses regarding egocentric interactions.
\end{itemize}

\section{Related Work}
\label{sec:formatting}
This paper mainly focuses on interaction understanding from an egocentric perspective and leverages the MLLMs to establish the ANNEXE model. Therefore, we elaborate the related work from three aspects: egocentric interaction understanding, vision-language foundation models, and multi-model egocentric understanding.

\textbf{Egocentric interaction understanding.}
Some existing works perform action recognition \cite{wang2023ego,li2022egocentric} to understand egocentric interactions coarsely, i.e., classify verbs and nouns of interactions in egocentric videos or images.
To improve recognition performance, action-relevant cues \cite{lee2020hand, li2015delving, zhou2016cascaded, kwon2021h2o, wang2020symbiotic, wang2020symbiotic1, shan2020understanding} such as hands, active objects, and their relations are considered to concentrate on executing interactions.
For example, Kown et al. \cite{kwon2021h2o} first proposed the H2O dataset and benchmark to predict hand pose and object pose for egocentric interaction recognition.
Wang et al. \cite{wang2020symbiotic, wang2020symbiotic1, shan2020understanding} utilized active object detection to concentrate more precisely on the occurring interaction.
Furthermore, Shiota et al. \cite{lu2021egocentric, shiota2024egocentric, yu2023fine, li2019deep, liu2021enhanced, wang2020symbiotic} considered the hand-object contact and object status to recognize interaction accurately.
Parallel to these, some other researchers utilized different modalities \cite{liu2020forecasting,sudhakaran2019lsta,shen2018egocentric, zatsarynna2021multi}, such as gaze \cite{huang2020mutual} and audio \cite{gong2023mmg}, to improve the recognition performance further.
For example, Wang \emph{et al.} \cite{wang2021interactive} leveraged the actor motion to improve the performance of classifying verbs and nouns.

Other recent existing works focus on detecting or segmenting the hands and interacting objects in an egocentric view. For example, Bambach et al. \cite{bambach2015lending} introduced a simple candidate region generation approach to locate and distinguish different hands and activities. Zhang et al. \cite{zhang2022fine,su2024ormnet} proposed to segment the active objects with the contacting hands. Leonardi et al. \cite{leonardi2022egocentric} proposed a method that detects hands and objects in the scene and determines which objects are currently involved in an interaction. 
\textbf{Remark}: These methods have completed the coarse understanding of egocentric interactions, but they lack a comprehensive understanding of interaction, like fluent text and fine-grained mask responses. In addition, the results cannot be directly applied to multiple downstream tasks with different requirements or queries.

\textbf{Vision-language foundation models.}
Recently, large language models (LLMs) \cite{ahn2022can,huang2022language,li2022pre,pasca2024summarize,brown2020language,ouyang2022training,wang2023openchat} have successfully achieved excellent performance on multiple language tasks due to their strong capabilities in understanding context. 
Consequently, vision-language models (VLMs) \cite{li2023blip,liu2024visual,zhu2023minigpt,awadalla2023openflamingo} have emerged by leveraging the advanced capabilities of large language models (LLMs). 
By combining linguistic and visual processing skills, VLMs can perform a range of tasks, such as visual question answering \cite{ning2023video,li2024mvbench,patraucean2024perception}, visual reasoning \cite{lu2019vilbert,radford2021learning,li2019visualbert}, and image retrieval \cite{yang2024ldre}. These models are capable of generating natural language descriptions based on input images, responding to image-related questions, and retrieving relevant images based on textual queries. 

MLLMs have also been widely used in visual grounding, which aims to identify objects in images according to given language descriptions. For example, Rasheed et al. \cite{rasheed2024glamm} proposed GLaMM to produce natural language responses intricately linked with corresponding object segmentation masks.
By combining the language generation capabilities of MLLMs with segmentation, Lai et al. \cite{lai2024lisa} demonstrated a large language-instructed segmentation assistant, LISA, that can manage complex reasoning tasks.
In addition, Yang et al. \cite{yang2022lavt} created the LAVT to integrate linguistic and visual features in the middle layers of a vision Transformer encoder, enhancing cross-modal alignment and producing precise segmentation outcomes with a lightweight mask predictor.

\begin{figure*}[ht]
	\centering
	\includegraphics[width=\textwidth]{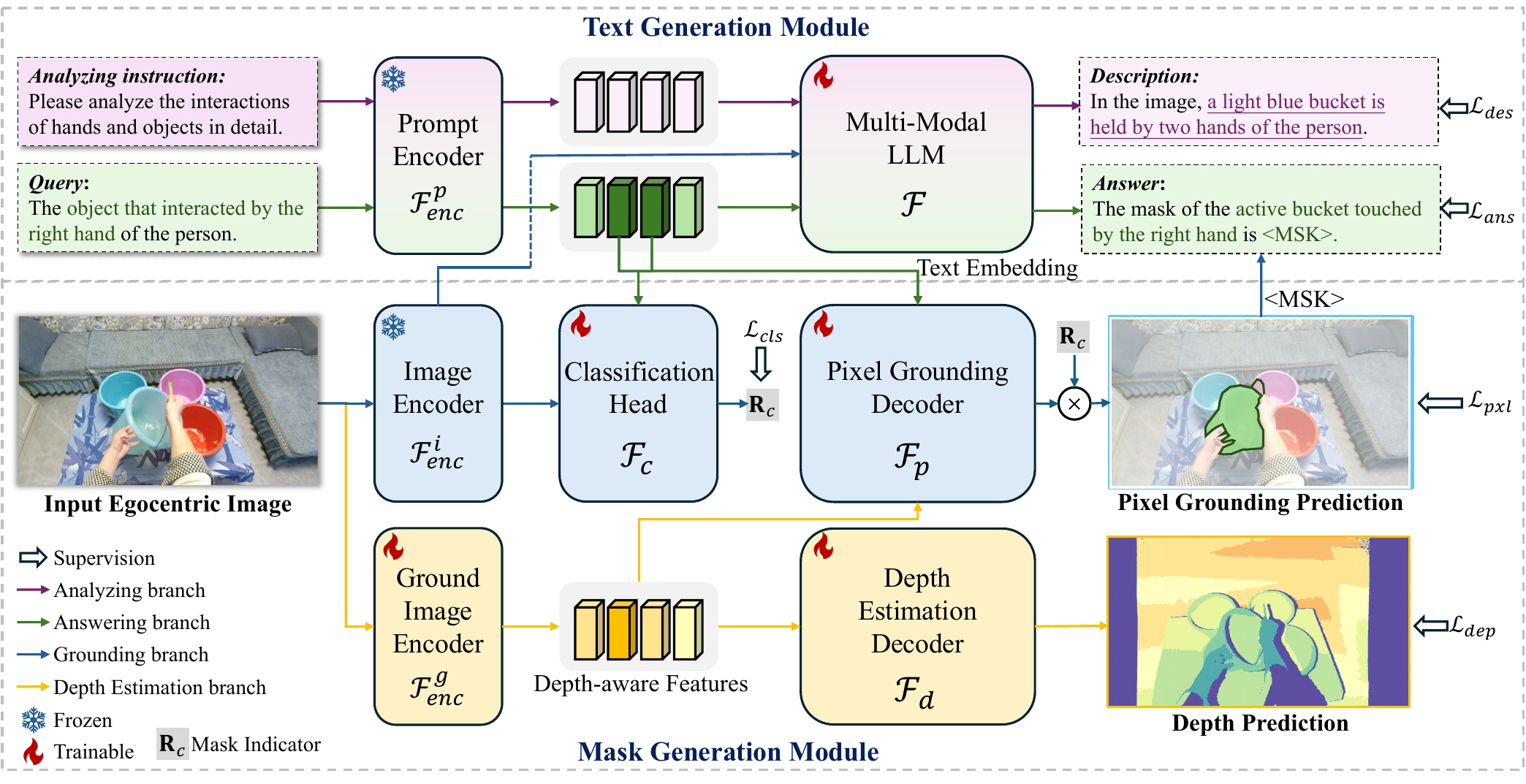}
	\caption{Overall architecture of the proposed ANNEXE model, a synergy of text generation and mask generation modules.}
	\label{fig:2}
    \vspace{-0.2cm}
\end{figure*}

\textbf{Multi-model egocentric understanding.}
Previous works have also attempted to combine visual and language information using MLLMs from an egocentric perspective.
Specifically, some researchers concentrated on egocentric video-language pretraining. For example, Lin et al. \cite{lin2022egocentric} created the first video-text pretraining EgoClip dataset containing the clip-text pairs. Based on this dataset, Valdez et al. \cite{valdez2024svitt} presented SViTT-Ego to solve the memory bottleneck problem existing in pertaining.

In addition, some people comprehended the content in egocentric images and videos through understanding and answering textual questions. 
For example, Jia et al. \cite{jia2022egotaskqa, di2024grounded,barmann2022did,chen2024grounded} carefully crafted questions to comprehend human actions and intents and proposed egoTaskQA to support the model in responding to these questions in egocentric videos.
Chen et al. \cite{chen2024grounded} proposed the multi-hop video question-answering task to provide temporal supporting evidence for further reasoning.
A small number of studies also combined egocentric action or interaction with VLM. For example, Kim et al. \cite{kim2025palm} proposed the PALM, which uses VLM to generate narrations and action sequences of past videos and predict future action sequences based on the generated content. In addition, Cheng et al. \cite{cheng2024egothink} proposed EgoThink to infer six tasks, including activity and localization.
The most relevant work with this paper is HOI-ref \cite{bansal2024hoi}, which uses language prompts to infer relevant content using bounding boxes regarding interaction. However, this work ignores analyzing the interaction and fails to achieve natural and fluent language output and pixel-level mask responses.

\textbf{Remark}. Although these methods have achieved coarsely visual grounding (bounding boxed) and/or text responses (verbs, nouns) for egocentric interactions, two problems still remain. First, they cannot generate specific outputs according to different queries, which limits the applications of egocentric interaction understanding in a variety of downstream tasks with distinct requirements.
Second, existing works fail to offer a thorough interpretation since they cannot incorporate text- and pixel-level responses to understand egocentric interactions.

\section{Methodology}
\label{methods}

This paper presents a novel task called Ego-IRG, which aims to generate both the text- and pixel-level responses to the user query about the interaction depicted in the egocentric image. 
The Ego-IRG task consists of three progressive sub-tasks, i.e., analyzing, answering, and pixel grounding.
We begin this section by formally defining the Ego-IRG task in Sec. \ref{sec::method-taskD}. 
In Sec. \ref{sec::method-OverA}, we introduce the proposed model ANNEXE in detail. 
Finally, the illustration of the proposed Ego-IRGBench is demonstrated in Sec. \ref{sec::method-dataset}.

\subsection{Task Definition}
\label{sec::method-taskD}
To facilitate the study of human actions, this paper proposes a novel Ego-IRG task, which unifies analyzing, answering, and pixel grounding regarding interactions in egocentric images based on user queries.
Taking an egocentric image $\textbf{I}$ and a query $\textbf{T}$ as input, the primary goal of the Ego-IRG task is to analyze the interaction in $\textbf{I}$ and produce the interaction description $\textbf{R}_{D}$.
Subsequently, the Ego-IRG task involves generating the answers $\textbf{R}_{A}$ containing the objects that need to be segmented in response to the query, along with the pixel-level mask grounding results $\textbf{R}_{M}$ of the object participating in the interaction specified in the query.

\subsection{ANNEXE Architecture}
\label{sec::method-OverA}
To address the proposed Ego-IRG task, this paper establishes a unified ANNEXE model, the first model that can generate text-level and fine-grained pixel-level responses simultaneously for egocentric interaction understanding.
As depicted in Fig. \ref{fig:2}, the ANNEXE model can be divided into two modules: text generation and mask generation. 
Specifically, taking an egocentric image $\textbf{I}$ and a query $\textbf{T}$ as input, the text generation module is used to generate the textual responses, including the interaction description $\textbf{R}_D$ and answers $\textbf{R}_A$ for the query.
Moreover, the mask generation module is designed to predict the fine-grained pixel-level grounding results $\textbf{R}_M$ according to the query, which shows the location and boundary for the referring objects.
Effectively integrating two modules, the ANNEXE model can complete the analyzing, answering, and pixel-grounding sub-tasks for comprehensive egocentric interactions. These two modules will deliberated as follows.

\subsubsection{Text Generation Module}
As illustrated in Fig. \ref{fig:2}, the text generation module mainly focuses on tackling two sub-tasks of Ego-IRG: analyzing and answering, which requires generating the interaction description and answers of the query.
Due to the significant success of the MLLMs, we utilize their strong capabilities to understand and align the context of language and image in the text generation module to predict text responses.
A typical approach is to use the prompt and image encoders to extract the textual and visual features of queries and images for prediction.
However, in our case, the analyzing sub-task lacks an explicit text prompt, which affects the confidence of generated interaction descriptions.
Hence, we introduce a hidden analyzing instruction $\textbf{T}_a$ to guide the MLLM to predict interaction description $\textbf{R}_D$ precisely.
Specifically, taking the query $\textbf{T}$ and hidden instruction $\textbf{T}_a$ as inputs, we employ a frozen prompt encoder \cite{chen2023minigpt} to extract their features. 
In addition, we use a frozen image encoder \cite{DBLP:conf/iclr/DosovitskiyB0WZ21} to extract the representative features of the input egocentric image $\textbf{I}$.
The process of extracting the features can be represented as:
\begin{align}
	\textbf{F}_{enc}^{img} &= \mathcal{F}_{enc}^i(\textbf{I},W_{enc}^i), \\
	\textbf{F}_{enc}^{ins}, \textbf{F}_{enc}^{que} &= \mathcal{F}_{enc}^p(\textbf{T}_a,\textbf{T},W_{enc}^p),
	\label{eq:extract}
\end{align}
where the $\textbf{F}_{enc}^{img}$, $\textbf{F}_{enc}^{ins}$, and $\textbf{F}_{enc}^{que}$ are the extracted features of the egocentric image, analyzing instruction, and input query, respectively.
And the functions $\mathcal{F}_{enc}^i(\cdot)$ and $\mathcal{F}_{enc}^p(\cdot)$ represent the image encoder with the parameters of $W_{enc}^i$ and the prompt encoder with the parameters of $W_{enc}^p$ respectively.

After feature extraction, the trainable multi-modal LLM $\mathcal{F}$ is utilized to obtain two types of textual output, \emph{i.e.}, interaction descriptions, and answers for the query about objects that need to be segmented.
This process can be represented as:
\begin{equation}
	\textbf{R}_{D}, \textbf{R}_{A} = \mathcal{F}(\textbf{F}_{enc}^{img}, \textbf{F}_{enc}^{ins}, \textbf{F}_{enc}^{que},W),
\end{equation}
where the $\mathcal{F}(\cdot)$ means the multi-modal LLM with the parameter of $W$.

In practice, the MLLM we used is Mini-GPT4v2 \cite{chen2023minigpt}, and the image and prompt encoders are frozen while the $\mathcal{F}$ is trainable.
Through this text generation module, the MLLM can be trained to generate coherent and precise interaction descriptions and answers depending on different queries and instructions, which further enhances the capability of the model to understand the in-depth information of input texts and images.

\begin{figure*}[t]
	\centering
	\includegraphics[width=\textwidth]{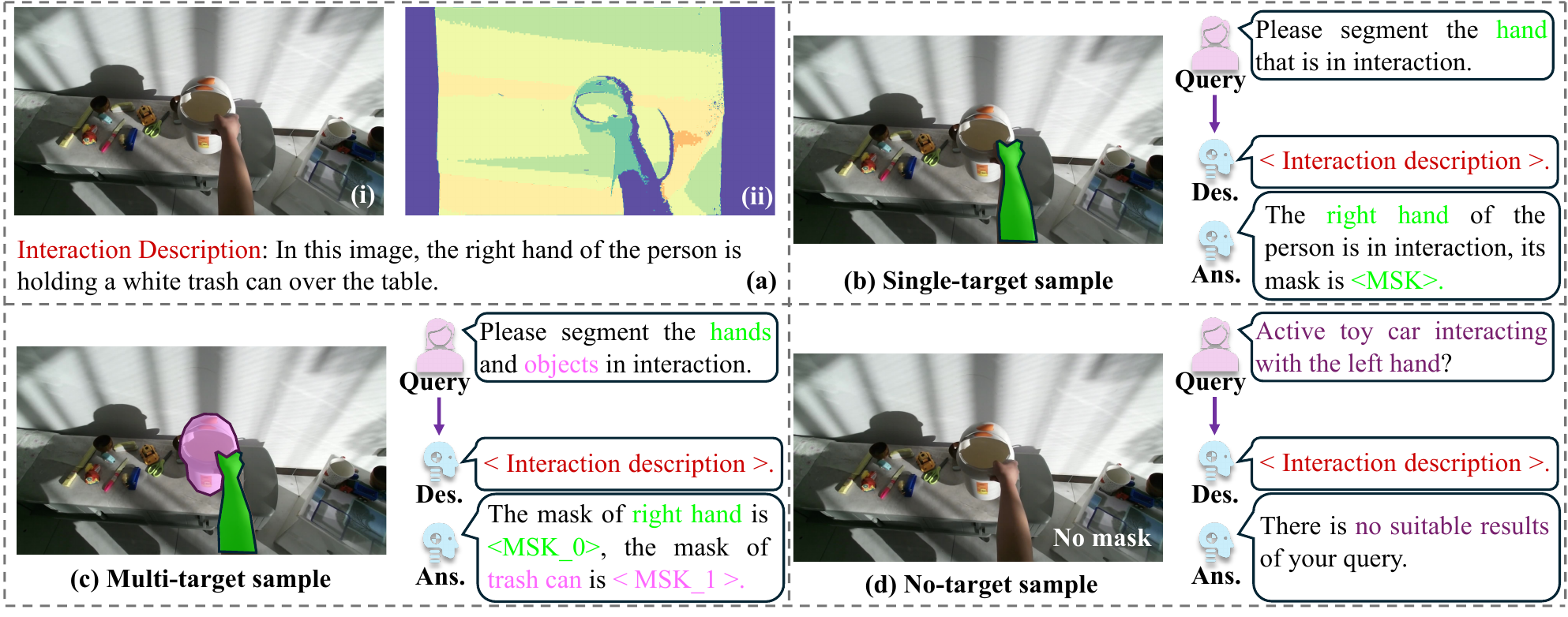}
	\caption{Illustration of the proposed Ego-IRGBench dataset, which includes (a)(i) the egocentric image, (a)(ii) the depth map, and interaction description. It also includes (b) single-, (c) multi-, and (d) no-target samples with the corresponding query, description (Des.), and answer (Ans.).}
	\label{fig:3}
    \vspace{-0.5cm}
\end{figure*}

\subsubsection{Mask Generation Module}
\label{sec:method:groundmodule}
The mask generation module is incorporated to complete the pixel grounding sub-task, which requires obtaining fine-grained masks of queried objects in the egocentric image.
To predict masks accurately, we observed that depth information is essential for pixel grounding about egocentric interactions. It is because when people perform an activity, the operating objects and hands are usually positioned in the foreground, which has a relatively clear distinction from the background in depth, especially when multiple similar objects are in the same scenario.
Therefore, we regard the depth estimation as an auxiliary task to integrate the depth information into the network.
As shown in Fig. \ref{fig:2}, the depth estimation branch is also incorporated to help distinguish the foreground and background for pixel grounding in the mask generation module.

Specifically, taking the image feature $\textbf{F}_{enc}^{img}$ and the query feature $\textbf{F}_{enc}^{que}$ as the input, we first use a trainable classification head $\mathcal{F}_c$ to predict the mask indicator, which indicates whether the targeted objects are in the image.
In this binary classification setting, if the predicted mask indicator is true, the pixel-level grounding mask will be generated further, while the mask prediction will not be considered if the mask indicator is false.
In addition, we select some representative features $\textbf{F}_{sel}^{que}$ from the query feature $\textbf{F}_{enc}^{que}$ along with $\textbf{F}_{enc}^{img}$ as inputs to the classification head $\mathcal{F}_c$.
In practice, the classification head includes several stacked convolution-relu-pooling layers with multi-layer perceptron.
The classification process is:
\begin{equation}
	\textbf{R}_c = \mathcal{F}_c(\textbf{F}_{enc}^{img},\textbf{F}_{sel}^{que}, W_c),
\end{equation}
where the $\textbf{R}_c$ is the predicted mask indicator with the value of 0 or 1, which is the intermediate output of the mask generation module. 
The $\mathcal{F}_{c}(\cdot)$ means the classification head with the parameters of $W_c$.

Furthermore, we regard depth estimation as an auxiliary task that helps the network extract image features that integrate latent deep information.
Specifically, we add an additional depth estimation branch in the mask generation module, in which a trainable ground image encoder $\mathcal{F}_{enc}^g$ is used to extract depth-aware image features $\textbf{F}_g^{img}$.
A trainable depth estimation decoder $\mathcal{F}_d$ with the parameters of $W_d$ is also utilized to predict the depth map of the input egocentric image, which encourages the ground image encoder to learn spatial information rather than appearance characteristics only.
The process of depth estimation is as follows:
\begin{align}
	\textbf{F}_g^{img} = \mathcal{F}_{enc}^g(\textbf{I}, W_g), \\
	\textbf{R}_d = \mathcal{F}_d(\textbf{F}_g^{img},W_d),
\end{align}
where the $\textbf{R}_d$ is the depth map prediction. And the $W_g$ is the parameter set of the ground image encoder $\mathcal{F}_{enc}^g$. In practice, we use the decoder in Depth Anything V2 \cite{yang2024depth} as the depth estimation decoder to predict the depth map.

Consequently, based on the selected query features $\textbf{F}_{sel}^{que}$, the extracted ground image features $\textbf{F}_g^{img}$, and the predicted mask indicator $\textbf{R}_c$, the predicted mask will be generated through the trainable pixel grounding decoder $\mathcal{F}_p$.
Specifically, taking $\textbf{F}_{sel}^{que}$ and $\textbf{F}_g^{img}$ as input, some convolutional and max-pooling layers are used to downsample these features into the same dimension. 
Then we add the $\textbf{F}_{sel}^{que}$ and $\textbf{F}_g^{img}$ with the same dimension to be the combined feature, which is sent into pixel grounding decoder $\mathcal{F}_p$ to predict the initial mask.
Finally, the initial mask is multiplied with the predicted mask indicator $\textbf{R}_c$ to obtain the final mask.
This process can be denoted as follows:
\begin{equation}
	\textbf{R}_M = \mathcal{F}_p((f_{down}(\textbf{F}_{g}^{img})+f_{down}^\prime(\textbf{F}_{sel}^{que})),W_p) \times \textbf{R}_c,
\end{equation}
where the $\textbf{R}_M$ is the generated mask, the $W_p$ is the trainable parameter of the pixel grounding decoder, the $\times$ means the multiplication of mask indicator and initial mask, and the $f_{down}(\cdot)$ and $f_{down}^\prime(\cdot)$ are the down-sampling functions.

By incorporating the mask generation module, the proposed ANNEXE model can predict precise and query-orientated pixel-level masks regarding egocentric interactions, which can be applied to various downstream tasks with different requirements more efficiently.

\subsubsection{Training}
\label{sec:method:train}
The training loss of the ANNEXE model consists of five parts, \emph{i.e.}, the analyzing loss $\mathcal{L}_{des}$ for interaction description, the answering loss $\mathcal{L}_{ans}$ for the response to queries, the classification loss $\mathcal{L}_{cls}$ for mask indicator, the depth estimation loss $\mathcal{L}_{dep}$, and the pixel grounding loss $\mathcal{L}_{pxl}$ for final mask.
We use the cross entropy loss for $\mathcal{L}_{des}$, $\mathcal{L}_{ans}$ \cite{zhu2023minigpt}, and $\mathcal{L}_{pxl}$.
The binary cross entropy loss is used for $\mathcal{L}_{cls}$, and smooth L1 loss is utilized for $\mathcal{L}_{dep}$.
Therefore, the overall loss function for the ANNEXE is:
\begin{align}
	\mathcal{L} &= \lambda_{des} \mathcal{L}_{des} + \lambda_{ans} \mathcal{L}_{ans} \\
	& + \lambda_{cls} \mathcal{L}_{cls}+	 \lambda_{dep} \mathcal{L}_{dep} + \lambda_{pxl} \mathcal{L}_{pxl},
\end{align}
where the $\mathcal{L}$ is the final loss, and the $\lambda_{des}$, $\lambda_{ans}$, $\lambda_{cls}$, $\lambda_{dep}$, and $\lambda_{pxl}$ are the weight of the corresponding loss function.

\begin{table}[]
\caption{Ego-IRGBench versus other datasets. Obviously, Ego-IRGBench is the first dataset that provides matching descriptions (Des.), answers (Ans.), and masks for the egocentric image-query inputs regarding interactions (Inter.).}
	\normalsize
	\resizebox{\linewidth}{!}{
		\begin{tabular}{@{}lcccccc@{}}
			\hline
\multirow{2}{*}{Dataset} & 
\multirow{1}{*}{Ego/} & \multirow{2}{*}{Inter.} & \multirow{2}{*}{Query} & \multicolumn{3}{c}{Responses} \\ \cline{5-7} 
& Exo  &   &  & Des. & Ans. & Mask \\ \hline
SAM-1B \cite{kirillov2023segment} & Exo &\textcolor{red}{\(\times\)} &\textcolor{green}{\checkmark} & \textcolor{red}{\(\times\)} &\textcolor{red}{\(\times\)}  &\textcolor{green}{\checkmark} \\ \hline
GranD \cite{rasheed2024glamm} & Exo & \textcolor{red}{\(\times\)} & \textcolor{red}{\(\times\)} & \textcolor{green}{\checkmark} & \textcolor{red}{\(\times\)} & \textcolor{green}{\checkmark} \\ \hline
RefCOCO \cite{kazemzadeh2014referitgame} &Exo &\textcolor{red}{\(\times\)} &\textcolor{green}{\checkmark} &\textcolor{red}{\(\times\)} &\textcolor{red}{\(\times\)} &\textcolor{green}{\checkmark} \\ \hline
RefCOCO+ \cite{kazemzadeh2014referitgame}&Exo &\textcolor{red}{\(\times\)} &\textcolor{green}{\checkmark} &\textcolor{red}{\(\times\)} &\textcolor{red}{\(\times\)} &\textcolor{green}{\checkmark} \\ \hline
RefClef \cite{kazemzadeh2014referitgame}&Exo &\textcolor{red}{\(\times\)} &\textcolor{green}{\checkmark} &\textcolor{red}{\(\times\)} &\textcolor{red}{\(\times\)} &\textcolor{green}{\checkmark} \\ \hline
G-Ref \cite{nagaraja2016modeling}&Exo &\textcolor{red}{\(\times\)} &\textcolor{green}{\checkmark} &\textcolor{red}{\(\times\)} &\textcolor{red}{\(\times\)} &\textcolor{green}{\checkmark} \\ \hline
VISOR \cite{VISOR2022}& Ego & \textcolor{green}{\checkmark} &\textcolor{green}{\checkmark} & \textcolor{red}{\(\times\)} & \textcolor{green}{\checkmark} &\textcolor{green}{\checkmark} \\ \hline
EPIC-KITCHENS \cite{Damen2022RESCALING}& Ego & \textcolor{green}{\checkmark} & \textcolor{red}{\(\times\)} & \textcolor{green}{\checkmark} & \textcolor{red}{\(\times\)} &\textcolor{red}{\(\times\)}\\ \hline
EGTEA \cite{li2018eye} & Ego & \textcolor{green}{\checkmark} & \textcolor{red}{\(\times\)} & \textcolor{green}{\checkmark} & \textcolor{red}{\(\times\)} & \textcolor{green}{\checkmark}\\ \hline
HOI4D \cite{liu2022hoi4d}& Ego & \textcolor{green}{\checkmark} & \textcolor{red}{\(\times\)} & \textcolor{red}{\(\times\)} & \textcolor{red}{\(\times\)} &\textcolor{green}{\checkmark} \\ \hline
Ego4D \cite{grauman2022ego4d} &Ego &\textcolor{green}{\checkmark} &\textcolor{green}{\checkmark} &\textcolor{red}{\(\times\)} &\textcolor{red}{\(\times\)} & \textcolor{red}{\(\times\)} \\ \hline
EgoHOS \cite{zhang2022fine}& Ego & \textcolor{green}{\checkmark} & \textcolor{red}{\(\times\)} & \textcolor{red}{\(\times\)} & \textcolor{red}{\(\times\)} &\textcolor{green}{\checkmark} \\ \hline
VISOR-NVOS \cite{shen2024learning}& Ego & \textcolor{green}{\checkmark} & \textcolor{red}{\(\times\)} & \textcolor{green}{\checkmark} &\textcolor{red}{\(\times\)} &\textcolor{green}{\checkmark}\\ \hline
\textbf{Ego-IRGBench} & Ego  & \textcolor{green}{\checkmark}  & \textcolor{green}{\checkmark} & \textcolor{green}{\checkmark} & \textcolor{green}{\checkmark}   &\textcolor{green}{\checkmark}  \\ \hline
\end{tabular}}

	\label{tab:1}
\vspace{-0.5cm}
\end{table}

\subsection{Ego-IRGBench}
\label{sec::method-dataset}
We develop a new Ego-IRG task to provide a comprehensive framework for comprehending egocentric interactions, as shown in Sec. \ref{sec::method-taskD}.
The objective of the Ego-IRG task is to generate the \textit{interaction descriptions, answers, and pixel-level masks} simultaneously based on the input text queries and egocentric images.
However, to the best of our knowledge, very few datasets can satisfy the requirements of Ego-IRG for the community to build up data-driven algorithms.
Therefore, to encourage future studies of the Ego-IRG task, we present a high-quality Ego-IRGBench dataset.
This dataset is built upon the HOI4D \cite{liu2022hoi4d} dataset, and we use a stage-wise annotation pipeline that consumes extensive manual efforts to re-label the dataset with queries and matching text- and pixel-level responses.
Unlike other related datasets, the Ego-IRGBench is the first to provide fluent textual responses and pixel-level grounding masks for understanding egocentric interaction. Tab. \ref{tab:1} displays the comparison of the Ego-IRGBench with other datasets.

\textbf{Structure and scale.} 
The Ego-IRGBench is a comprehensive dataset that includes over 1.6 million queries along with their corresponding text- and pixel-level response pairs. This extensive dataset is built upon a collection of 20,504 RGB-D egocentric image pairs extracted from the HOI4D dataset \cite{liu2022hoi4d}, including various interactions and environments from a first-person view. 
Specifically, each egocentric RGB image is paired with a depth map (as shown in Fig. \ref{fig:3} (a)(ii)) and an interaction description (Fig. \ref{fig:3}), providing spatial information about the scene and outlining the specific interaction taking place.
Furthermore, multiple queries are labeled for each image, allowing for various inquiries related to the interactions depicted. Each query is paired with an answer and corresponding pixel-level grounding mask, forming systematic and ultimate feedback to the query. 
It is worth emphasizing that the query is not limited to inferring a single target (Fig. \ref{fig:3} (b)) involving interactions. Multi-target (Fig. \ref{fig:3} (c)) and no-target (Fig. \ref{fig:3} (d)) queries are also provided in the Ego-IRGBench dataset, which shows the diversity of the dataset. 
More explanations for the Ego-IRGBench dataset can be found in Appendix A.1.

\textbf{Step-wise annotation pipeline.}
To efficiently build up the dataset that meets the requirements of the Ego-IRG task, we implemented a semi-automated annotation pipeline from coarse to fine. 
Specifically, the annotation pipeline includes three steps: interaction classification, hand-object mask generation, and query-response generation.
In the first step, the employed experts are asked to classify the aligned RGB-D images to determine whether interactions are taking place. As a result, the images are divided into two categories: with and w/o interactions. 
For samples that include interactions, the description of the interaction is generated semi-manually. 
In the second step, masks for the hands and interacting objects are generated semi-automatically based on the masks of the HOI4D \cite{liu2022hoi4d} dataset.
Finally, queries and textual responses are created manually, while the corresponding pixel-level grounding masks are generated automatically.
More details about annotations are explained in Appendix A.2.

\textbf{Evaluation criteria.} To build up the benchmark for Ego-IRG, we split the Ego-IRGBench into train, validation, and test sets according using a ratio of 5:2:3, yielding a train set of 10,249 images with 806,982 queries, a validation set of 4,094 images with 322,208 query-response-mask pairs, and a test set of 6,161 images and 485,174 query-response-mask pairs. 
In validation and testing, three key aspects are evaluated: i) generated
interaction description quality, ii) quality of generated answers for queries, and iii) pixel-level grounding accuracy.
To examine the qualities of generated interaction descriptions and answers, METEOR \cite{banerjee2005meteor} and CIDEr \cite{vedantam2015cider} metrics are considered following previous work \cite{rasheed2024glamm}.
To verify the accuracy of pixel grounding, cIoU is considered as the metric following the work in \cite{lai2024lisa,xia2024gsva}.
More details are illustrated in Appendix A.3.

\vspace{-0.2cm}
\section{Experiments}
\label{Experiments}
This paper proposes a unified Ego-IRG task for egocentric interaction understanding, along with the Ego-IRGBench dataset and ANNEXE model. 
We first provide the dataset that we used for the experiments in this section, followed by implementation details in Sec. \ref{sec:exper:dataset}.
The effectiveness of the suggested ANNEXE is then demonstrated by the quantitative results, which are shown in Sec. \ref{sec:exper:quan}.
Finally, the qualitative results are displayed in Sec. \ref{sec:exper:qual} to visualize the results of ANNEXE for the Ego-IRG task.
\vspace{-0.2cm}
\subsection{Dataset and Implementation Details}
\label{sec:exper:dataset}
As explained in Sec. \ref{sec::method-dataset}, we proposed a comprehensive Ego-IRGBench dataset that consists of over 1.6 million query-response-mask pairs about egocentric interactions. 
This dataset is used for training and testing.

For network architecture, unless otherwise specified, we use Mini-GPT4v2 \cite{chen2023minigpt} as the base multi-modal LLM $\mathcal{F}$, and adopt the Depth Anything V2 \cite{yang2024depth} as the depth estimation branch. 
Additionally, we utilize the instruction template {$ [INST] \; <Img><ImageHere></Img> \; [task] \; Instruction \; [/INST]$} as the template of input query and analysis instruction.
The experiments are conducted on 4 NVIDIA 6000Ada GPUs.
During training, the batch size per GPU is set to 4, and we use the AdamW optimizer with the learning rate and weight decay set to 1e-5 and 0.05, respectively.
We also adopt linear WarmupCosineLR as the learning rate scheduler, setting the warmup iterations to 1000.
For image preprocessing, the image is resized to (448, 448) during training and testing and normalized with the mean and variant of [0.55,0.55,0.53] and [0.22,0.23,0.25].

\vspace{-0.2cm}
\subsection{Quantitative Results}
\label{sec:exper:quan}

In this section, we first compare the results of the proposed ANNEXE with other methods to demonstrate its effectiveness. Then, we conduct ablation experiments on the depth estimation branch to indicate the benefit of this auxiliary task for mask grounding. Finally, we set different hyperparameters for comparison to show the influences of various loss weights.

\subsubsection{Comparison Results}
To validate the effectiveness of the proposed ANNEXE model in understanding the egocentric interactions, we conduct comparison results with other methods on the Ego-IRGBench dataset.
However, to the best of our knowledge, few methods can accomplish analyzing, answering, and pixel grounding in a unified manner.
Therefore, we compare our method with other similar methods that can achieve partial functionality.
Specifically, we first compare the interaction analyzing and answering abilities of the ANNEXE with some image captioning methods, which can generate descriptions of happening interactions according to instructions and queries.
Furthermore, we compare the pixel-level grounding ability of ANNEXE with some referring image segmentation methods, which can generate mask grounding results based on the query.

\textbf{Interaction analyzing and answering results.}
The first and second sub-tasks of the proposed Ego-IRG task are to generate descriptions of interactions between hands and objects and answer what needs to be segmented according to the query, which is similar to the image captioning task given different text prompts. 
Therefore, we selected some models in image captioning to validate the effectiveness of the interaction analyzing and answering abilities of ANNEXE. 
The comparison results of analyzing and answering results are shown in Tab. \ref{tab:2} and Tab. \ref{tab:2.5}, respectively.
Our proposed ANNEXE model achieves the best results even compared with other LLM-based methods such as SmallCap \cite{ramos2023smallcap} and PromptCap \cite{hu2023promptcap}, which can generate fluent and coherent descriptions and answers about the egocentric interactions taking place according to the query.

\begin{table}[t]
\caption{ANNEXE versus other methods about analyzing sub-task on the Ego-IRGBench validation and test sets. The metrics used are METEOR and CIDEr.}
\normalsize
\centering
\resizebox{\linewidth}{!}{
\begin{tabular}{l|cc|cc}
\hline
\multirow{2}{*}{Model} 
& \multicolumn{2}{c}{Val}              
& \multicolumn{2}{c}{Test} \\ 
& M $\uparrow$ & C $\uparrow$ & M $\uparrow$  & C $\uparrow$  \\ \hline
ViECap (ICCV 2023) \cite{fei2023transferable}
& 0.063  & 0.096   & 0.096 & 0.094   \\ \hline
PromptCap (ICCV 2023) \cite{hu2023promptcap} 
& 0.147 &0.176 &0.148 &0.184 \\ \hline
SmallCap (CVPR 2023) \cite{ramos2023smallcap} 
&0.177 & 0.219  & 0.179 & 0.221   \\ \hline
ConZIC (CVPR 2023) \cite{zeng2023conzic} &0.102&0.063&0.102&0.067 \\ \hline
\rowcolor{gray!30} \textbf{Ego-IRG}&\textbf{0.563}&\textbf{1.516} &\textbf{0.563} &\textbf{1.494}                       \\ \hline
\end{tabular}}
\label{tab:2}
\end{table}

\begin{table}[t]
\caption{ANNEXE versus other methods about answering sub-task on the Ego-IRGBench validation and test sets. The metrics used are METEOR and CIDEr.}
\normalsize
\centering
\resizebox{\linewidth}{!}{
\begin{tabular}{l|cc|cc}
\hline
\multirow{2}{*}{Model} 
& \multicolumn{2}{c}{Val}              
& \multicolumn{2}{c}{Test} \\ 
& M $\uparrow$ & C $\uparrow$ & M $\uparrow$  & C $\uparrow$  \\ \hline
ViECap (ICCV 2023) \cite{fei2023transferable}
& 0.019  & 0.024   & 0.021 & 0.026   \\ \hline
SmallCap (CVPR 2023) \cite{ramos2023smallcap} 
&0.034 & 0.091  & 0.035 & 0.092   \\ \hline
PromptCap (ICCV 2023) \cite{hu2023promptcap} 
& 0.086 &0.205 &0.089 &0.218 \\ \hline
\rowcolor{gray!30}\textbf{Ego-IRG}&\textbf{0.363}&\textbf{2.543} &\textbf{0.365} &\textbf{2.590}                       \\ \hline
\end{tabular}}
\label{tab:2.5}
\end{table}

\begin{table}[]
\caption{ANNEXE versus other referring image segmentation methods on the Ego-IRGBench validation and test sets to prove the pixel grounding capability. The metric for pixel-grounding is cIoU.}
\resizebox{\linewidth}{!}{
	\begin{tabular}{l|c|c}
		\hline
Model   & 
\multicolumn{1}{c|}{Val.cIoU $\uparrow$} & \multicolumn{1}{c}{Test.cIoU $\uparrow$} \\ \hline
GSVA (CVPR 2024) \cite{xia2024gsva}& 12.92 & 13.32   \\ \hline
GRES (CVPR 2023) \cite{liu2023gres}& 19.88   & 21.35    \\ \hline
LISA (CVPR 2024) \cite{lai2024lisa} &21.74&22.63 \\ \hline
X-Decoder (CVPR 2023) \cite{zou2023generalized}& 23.72    & 23.75 \\ \hline
\rowcolor{gray!30}\textbf{Ego-IRG} & \textbf{35.14} \textcolor{blue}{(+11.42)}    & \textbf{36.02} \textcolor{blue}{(+12.27)}   \\ \hline
\end{tabular}}
	\label{tab:3}
    \vspace{-0.5cm}
\end{table}

\textbf{Pixel grounding results.}
To validate the performance of the pixel grounding of ANNEXE, we compare it with some of the latest referring image segmentation methods (including LLM-based methods such as GSVA \cite{xia2024gsva} and LISA \cite{lai2024lisa}) on the validation and test sets of the Ego-IRGBench dataset. 
The results are shown in Tab. \ref{tab:3}, which shows that our ANNEXE achieves state-of-the-art segmentation performance in referring the objects involving interactions according to queries.
Therefore, the proposed ANNEXE model can efficiently understand the context of multi-modal inputs and perform fine-grained pixel grounding for various downstream tasks.

\begin{table}[]
\caption{Ablation study about the depth estimation branch.}
\centering
\resizebox{\linewidth}{!}{
\begin{tabular}{ccc|c|c}
\hline
\multicolumn{1}{c|}{} & \multicolumn{1}{c|}{Sub-task}   & \multicolumn{1}{c|}{Metrics}  & Without depth & With depth \\ \hline
\multicolumn{1}{c|}{\multirow{5}{*}{Val}}  & \multicolumn{1}{c|}{\multirow{2}{*}{Analyzing}} & METEOR  & 0.537  & 0.563      \\ \cline{3-5}
\multicolumn{1}{c|}{}                      & \multicolumn{1}{c|}{}                      & CIDEr   & 1.447  & 1.516      \\ \cline{2-5} 
\multicolumn{1}{c|}{}                      & \multicolumn{1}{c|}{\multirow{2}{*}{Answering}} & METEOR       & 0.352   & 0.363      \\ \cline{3-5}
\multicolumn{1}{c|}{}                      & \multicolumn{1}{c|}{}                      & CIDEr        & 2.312  & 2.543      \\ \cline{2-5} 
\multicolumn{1}{c|}{} & \multicolumn{1}{c|}{Grounding} & \multicolumn{1}{c|}{cIoU} 
&  31.72 & 35.14  \\ \cline{1-5}
\multicolumn{1}{c|}{\multirow{5}{*}{Test}} & \multicolumn{1}{c|}{\multirow{2}{*}{Analyzing}} & METEOR   & 0.536   & 0.563    \\ \cline{3-5}
\multicolumn{1}{c|}{}                      & \multicolumn{1}{c|}{}                      & CIDEr       & 1.423  & 1.494      \\ \cline{2-5} 
\multicolumn{1}{c|}{}                      & \multicolumn{1}{c|}{\multirow{2}{*}{Answering}} & METEOR  & 0.350  & 0.365     \\ \cline{3-5}
\multicolumn{1}{c|}{}                      & \multicolumn{1}{c|}{}                      & CIDEr       & 2.317   & 2.590    \\ \cline{2-5}
\multicolumn{1}{c|}{} & \multicolumn{1}{c|}{Grounding} & \multicolumn{1}{c|}{cIoU} 
 & 31.90 &36.02\\ \cline{1-5}
\end{tabular}}
\label{tab:4}
\end{table}

\subsubsection{Ablation Study}
As introduced in Sec. \ref{sec:method:groundmodule}, we integrate depth estimation as an auxiliary task for precise pixel grounding.
In this section, we conduct the ablation study for the depth estimation branch to prove its ability to distinguish the foreground and background. The results are shown in Tab. \ref{tab:4}.
As observed from the results, when the depth estimation is incorporated, the performance of pixel grounding will be improved by 3.42\% and 4.12\% on validation and test sets, respectively, which indicates the necessity of depth information.

\begin{table}[]
\caption{Hyper-parameter experiments of ANNEXE on the Ego-IRGBench test set. 'Ans.' means answering sub-task, 'Ana.' means the analyzing sub-task, 'M' and 'C' represent the metrics METEOR and CIDEr, and cIoU is the metric of pixel grounding.}
	\resizebox{\linewidth}{!}{
	\begin{tabular}{ccccc|ccccc}
		\hline
{$\lambda_{des}$} & 
{$\lambda_{ans}$} & 
{$\lambda_{pxl}$} & 
{$\lambda_{cls}$} & 
{$\lambda_{dep}$} & 
Ana.M                    
& Ana.C
& Ans.M
& Ans.C
& cIoU  \\ \hline
 1.0 & 1.0 & 1.0 & 1.0 & 1.0 
 & 0.538 & 1.421 & 0.346 & 2.328 & 23.34      \\ \hline
 0.5 & 0.5 & 0.5 & 1.0 & 1.0 
 & 0.523 & 1.410 & 0.352 & 2.316 & 22.46      \\ \hline
 \textbf{1.0} & \textbf{1.0} & \textbf{1.0} & \textbf{0.5} & \textbf{0.5} 
 & \textbf{0.563} & \textbf{1.494} & \textbf{0.365} & \textbf{2.590} & \textbf{36.02}      \\ \hline
	\end{tabular}}
\label{tab:5}
\vspace{-0.5cm}
\end{table}

\subsubsection{Hyper-parameter Study}
As described in sec. \ref{sec:method:train}, the weights $\lambda_{des}$, $\lambda_{ans}$, $\lambda_{cls}$, $\lambda_{dep}$, and $\lambda_{pxl}$  are needed to be set during training. 
In this section, we study the influences of different hyper-parameter settings.
The results are shown in Tab. \ref{tab:5}, which exhibits great differences between diverse settings.
Specifically, when the weights of analyzing, answering, and pixel grounding are smaller than the weights of indicator classification and depth estimation, the performance will be decreased. 
It is mainly because the interaction analyzing, answering, and pixel grounding are the primary goals of the proposed ANNEXE, and the depth estimation and index classification are designed to assist pixel grounding. 

\subsection{Qualitative Evaluation}

\label{sec:exper:qual}
Due to the page limitation, we show qualitative results for Ego-IRG task on the Ego-IRGBench dataset in Appendix A.4.

\vspace{-0.2cm}
\section{Conclusion}

To comprehensively interpret egocentric interactions, this paper proposes the novel Ego-IRG task, which encompasses analyzing, answering, and pixel grounding sub-tasks. Furthermore, this paper creates a large-scale Ego-IRGBench dataset, which includes egocentric images and queries with corresponding multi-modal responses. Finally, the ANNEXE model is established to generate the text- and pixel-level outputs precisely according to diverse queries.

\section{Acknowledgement}
The research work was conducted in the JC STEM Lab of Machine Learning and Computer Vision funded by The Hong Kong Jockey Club Charities Trust.

{
    \small
    \bibliographystyle{ieeenat_fullname}
    \bibliography{main}
}


\end{document}